# Unsupervised multi-branch Capsule for Hyperspectral and LiDAR classification

Xu Quanfeng [1], Tang Yi [1*] and She Yumei [1]

[1]School of Mathematics and Computer Science, Yunnan Minzu University, 2929 Yuehua Street, Kunming, 650500, China

**ABSTRACT**

With the convenient availability of remote sensing data, how to make models to interpret complex remote sensing data attracts wide attention. In remote sensing data, hyperspectral images contain spectral information and LiDAR contains elevation information. Hence, more explorations are warranted to better fuse the features of different source data. In this paper, we introduce semantic understanding to dynamically fuse data from two different sources, extract features of HSI and LiDAR through different capsule network branches and improve self-supervised loss and random rigid rotation in Canonical Capsule to a high-dimensional situation. Canonical Capsule computes the capsule decomposition of objects by permutation-equivariant attention and the process is self-supervised by training pairs of randomly rotated objects. After fusing the features of HSI and LiDAR with semantic understanding, the unsupervised extraction of spectral-spatial-elevation fusion features is achieved. With two real-world examples of HSI and LiDAR fused, the experimental results show that the proposed multi-branch high-dimensional canonical capsule algorithm can be effective for semantic understanding of HSI and LiDAR. It indicates that the model can extract HSI and LiDAR data features effectively as opposed to existing models for unsupervised extraction of multi-source RS data.

Capsule Networks; LiDAR; Remote Sensing; Dimensionality Reduction.

**1.Introduction**

Remote sensing (RS) technology can collect data with massive amount of ground information, including Hyperspectral imagery (HSI) and light detection and ranging (LiDAR). HSI records the spectral information carried by hundreds of continuous spectral bands, which reflects the physical information of the ground surface. Furthermore, LiDAR provides elevation information of the ground surface. The combination of HSI and LiDAR may be highly effective [1-3] and is widely used in studies such as crop delineation, urban delineation and land cover mapping [4-6].

To extract features from Remote Sensing images, the existing methods can be generally divided into three groups: spectral methods, spectral-spatial methods and spectral-spatial-elevation methods (HSI+LiDAR). The spectral methods exploited traditional machine learning models like random forests [7-8] and support vector machines (SVM) [9-10] to extract spectral features from spectral signatures. Subsequently, deep learning such as convolutional neural networks (CNN) [11-13] and capsule networks [14-16] showed their advantages. The spectral-spatial methods extracted spatial and spectral information with CNN. For example, Sun et al. [17] proposed Spectral-Spatial Attention Network (SSAN), in which attention was used to search the effective features of HSI cubes. Song's and Wang's researches [18-19] also revealed that spectral-spatial methods can better extract the information in HSI.

---

[*]Corresponding author: yitang.math@gmail.com

The spectral-spatial-elevation methods fuse HSI and LiDAR to classify features, where LiDAR [20-21] provides complementary altitude information for hyperspectral. For instance, Zhang et al. [6,22] proposed two-branch CNN to extract HSI and LiDAR information, respectively. Du et al. [23] adopted graph-embedding to fuse multi-source features of HSI and LiDAR. The tight fusion of HSI and LiDAR demonstrated comprehensive RS data features, proving more satisfying performance of HSI + LiDAR than that of HSI alone. The above work explores remote sensing data from spatial and spectral feature extraction. However, the existing methods rely heavily on the supervision of labels, which is unreasonable in practice. To solve this problem, we propose an unsupervised fusion method to simultaneously extract spectral, spatial and elevation information.

For the same Earth's surface category, the combination of spatial location and spectral information are complementary., In the boundary between different Earth's surface category, the complementary spatial location plays a negative role in the extraction of the boundary features. Canonical Capsules algorithm [24] effectively identifies the semantic distinction to each part of the data in the decomposition and is able to achieve effective semantic decomposition and reconstruction of the dataset. In this paper, the canonical capsule algorithm is exploited to extract the spatial location and spectral information of the small patch according to the semantic understanding.

The Canonical Capsules algorithm can't compute high-dimensional data. This paper transforms the input dimension of the model information extraction and the random rigid rotational transformation of data into the high-dimensional case, and conducts an unsupervised semantic understanding upon HSI and LiDAR. Then, Capsules Networks are utilized to extract spatial and spectral features. Based on the above issues, the main contributions of this paper are:

(1) To achieve an unsupervised semantic understanding upon HSI and LiDAR, the Canonical Capsules algorithm is used to where the input dimension of the 3D point cloud model and the random rigid rotation matrix of the point cloud are improved to high-dimension.

(2) A structure for extracting spectral-spatial-elevation information of HSI and LiDAR data is proposed. One branch of structure extracts preliminary features of HSI with a capsule network and conducts a semantic understanding upon these features with a high-dimensional canonical capsule network. The other branch extracts feature directly from 3D LiDAR data with canonical capsules. When the attention map distribution of HSI branches approximates the attention map distribution of LiDAR, the attention information fusion is realized in the process of multi-source feature learning.

## 2.Related works

In this section, the classification of HSI and LiDAR data, and the related 3D canonical capsule algorithms are introduced subsequently.

### 2.1 HSI and LiDAR classification

HSI and LiDAR classification describes pixel-level land cover mapping from different data sources. At present, it is a common practice to use CNN to extract the features of two data, respectively. For instance, Zhang et al. [6] proposed a complex CNN structure to extract HSI and LiDAR features and then concatenate them for classification. Zhao et al. [5] proposed hierarchical Random Walk network (HRWN), which uses pixelwise affinity branch to extract spatial information in LiDAR to assist HSI in feature extraction. Nowadays, there are plenty of novel ideas on the fusion of HSI and LiDAR. Jia et al. [25] employed Kernel Principal Component analysis (KPCA) to reduce the size of HSI as well as noise, and random forest decision was introduced to fuse HSI and LiDAR for classification. In Dong et al.'s research [26],

self- and cross-guided attention (SCGA) assigned higher weights to regions of interest and channels of LiDAR and HSI maps to realize attention-based fusion of HSI and LiDAR. Du et al. [23] carried out feature fusion of HSI and LiDAR with Graph Embedding in a single network, and superior classification effects were obtained.

### 2.2 Canonical Capsules

Sun et al. [24] proposed the Canonical Capsules algorithm, a completely unsupervised learning method based on points of 3D point clouds, which truly achieves completely unsupervised learning on large-scale datasets. This algorithm decomposes the point cloud by k-folds to obtain the primary capsules, and then used those primary capsules to estimate the invariant point clouds shape and the equivariant point clouds center of each part. By associating the capsule pose with the capsule shape (invariance), an object-centered coordinate framework is constructed to support the subsequent auto-encoders network. Its unsupervised learning on the ShapeNet dataset and accuracy outperform both the AtlasNetV2 network [27] and the 3D-PointCapsNet network [28].

The structure of canonical capsules algorithm is similar to the autoencoder. The attention map and feature map obtained during the encoding process jointly describe the semantic information of the point cloud. The model can obtain semantic segmentation unsupervised, and point features containing semantic information can also be obtained for further exploration.

The reconstructed point cloud and original point cloud chamfering loss could be obtained by decoding the invariant features and equivariant features. In the process of BP, the increasingly small loss enables the hidden variables to contain most of the information in the original point cloud and the point cloud to have correct semantic judgment.

## 3. The proposed method

HSI+LiDAR is adopted in this study. Firstly, two sets of patches, assumed as $X^H$ and $X^L$, are obtained based on HSI and LiDAR with consistent spatial resolution. Specifically, in the set of HSI $X^H = \{X_{11}^H, \cdots, X_{ij}^H, \cdots, X_{hw}^H\}, X_{ij}^H \in \mathbb{R}^{b \times b \times C}$, $X_{ij}^H$ represents a HSI patch centered at $ij$ with a patch size of $b \times b$. C is the number of spectral channels of HSI. Similarly, in the set $X^L$ of LiDAR, $X^L = \{X_{11}^L, \cdots, X_{ij}^L, \cdots, X_{hw}^L\}, X_{ij}^L \in \mathbb{R}^{b \times b \times 3}$, $X_{ij}^H$ represents a small LiDAR set centered at $ij$ and with a patch size of $b \times b$, and the last dimension contains the information of location and elevation.

High dimensional data mentioned above act as the object of semantic understanding, and the unsupervised understanding of local semantic information is conducted from the entire object. Specially, a given point cloud $P \in R^{X \times D}$ denoting x points in the D dimension make up these point clouds $P$, is transformed into K-fold attention map $A \in R^{X \times K}$ for K capsules and per-point feature map $F \in R^{X \times C}$ with C channels in encoder $\varepsilon$ using CNN and attentive context normalization (ACN) [29].

$$A, F = \varepsilon(P) \tag{1}$$

$$\theta_k^e = \frac{\sum_p A_{p,k}^e P_p^e}{\sum_p A_{p,k}^e}, \beta_k^e = \frac{\sum_p A_{p,k}^e F_p^e}{\sum_p A_{p,k}^e} \tag{2}$$

In this paper, we improve to perform only one high-dimensional rigid rotation $T^e$ on it, and transform it into $A^e, F^e = \varepsilon(T^e(P))$ with a high-dimensional encoder by ACN calculation. In equation 2, $\theta_k \in R^D$ denotes the pose of the $k^{th}$ capsule, which is described by the location information in D-dimensional space, and $\beta_k \in R^C$ correspond to capsule shape information. The position parameter $\theta_k$ has spatial equivariance and can be realized after rotation and translation. The $\beta_k$ denotes the shape information of the point clouds which is invariant. Then, $\theta_k$ and $\beta_k$ are used to calculate the loss of their invariance and equivariance, as in the following equation (3) and (4).

$$L_{equ} = \frac{1}{K}\sum_k \|(T^e)\theta_k - \theta_k^e\|_2^2 \tag{3}$$

$$L_{inv} = \frac{1}{K}\sum_k \|\beta_k - \beta_k^e\|_2^2 \quad (4)$$

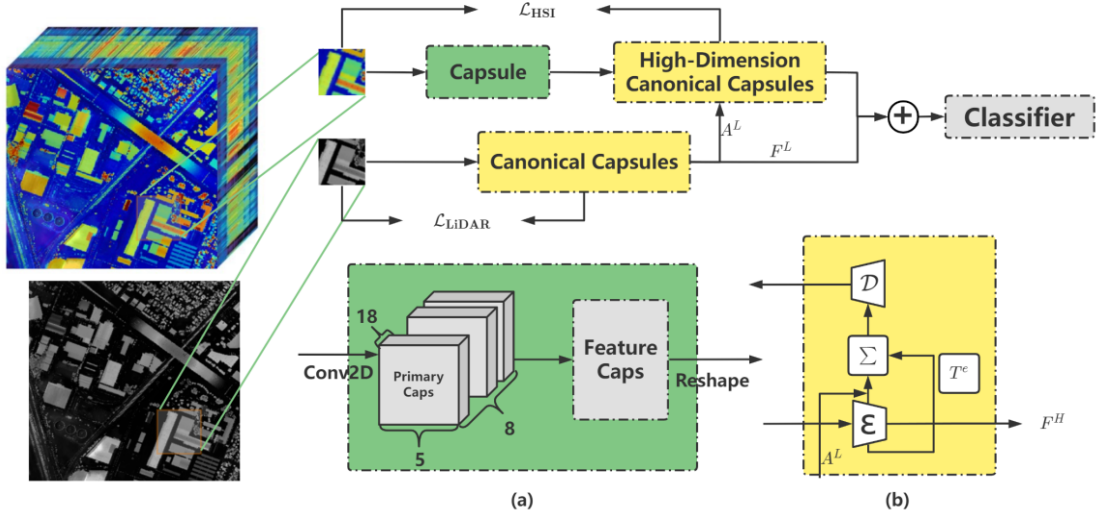

Fig 1. Unsupervised Capsule Network Structure

(a) Capsule Block (b) High-Dimension Canonical Capsules Block.

$$L_{HSI} = L_{H\_equ} + L_{H\_inv} + L_{H\_chamfer} \quad (5)$$
$$L_{LiDAR} = L_{L\_equ} + L_{L\_inv} + L_{L\_chamfer} \quad (6)$$
$$L_{H\_L} = KL(A^H, A^L) \quad (7)$$
$$L = \alpha \cdot L_{HSI} + \beta \cdot L_{LiDAR} + \gamma \cdot L_{H\_L} \quad (8)$$

Based on high dimensional canonical capsule algorithm, we use two branches to extract the spectral features in HSI and the spatial elevation information of LiDAR respectively. The specific structure is shown in Fig. 1. As for the branch extracting spectral information, it firstly extracts preliminary features of small patches of HSI data by capsule network. After semantic understanding of these features by high-dimensional canonical capsule, $A^H$ and $F^H$ are obtained, which respectively contain attention map and feature map of preliminary features. Then, the semantic understanding features are decoded and then approach original HSI through BP, and the HSI loss function is shown in equations 5. The branch for extracting spatial information directly adopts the canonical capsule model to extract the spatial elevation information of small patches of LiDAR to the attention map $A^L$ and the feature map $F^L$. In the process of two-branch learning, the distribution of attention map of constraining spectrum is close to that of spatial attention map, and KL divergence is used to evaluate the distribution of attention map, which is shown in the loss function 7. The summarized loss function is shown in Equation 8. In this paper, $\alpha = 0.5, \beta = 0.5$ and $\gamma = 0.1$ are adopted for the experiment. Finally, the features of $F^L$ containing spatial location information and $F^H$ containing spectral information are fused. Furthermore, in order to verify the unsupervised extraction effect of HSI and LiDAR features, SVM is used to classify semantic features.

## 4. Experiments

This section validates the effect of high-dimensional capsule model for semantic understanding of fused HSI and LiDAR data in MUUFL. Specifically, Section 4.1 introduces Houston 2013 datasets. Section 4.2 introduces comparative methods adopted in this study. In section 4.3, experiment settings and environment are verified. In Section 4.4, the features containing semantic understanding are extracted from the MUUFL and Houston 2013 dataset, and the model feature extraction effect is verified using a simple classifier SVM.

## 4.1 Datasets

This section describes the MUUFL and Houston 2013 datasets used in this paper and introduces how to construct the dataset for this paper and the amount of data for training and testing.

(a) *MUUFL Gulfport DataSet*: This dataset is obtained on the campus of University of Southern Mississippi Gulf Park, Long Beach Mississippi. The dataset contains HSI and LiDAR at the same spatial resolution, with 11 categories at an image size of 325×220 pixels. It originally contained 72 bands, but 64 appropriate bands are finally left after removing the eight bands affected by noise.

Table 1. Training and testing samples of MUUFL Dataset, with 5% of the per class for training

| No. | Class | Train | Test | All |
|---|---|---|---|---|
| 1 | Trees | 1162 | 22084 | 23246 |
| 2 | Grass Pure | 213 | 4057 | 4270 |
| 3 | Grass Groundsurface | 344 | 6538 | 6882 |
| 4 | Dirt and Sand | 91 | 1735 | 1826 |
| 5 | Road Materials | 334 | 6353 | 6687 |
| 6 | Water | 23 | 443 | 466 |
| 7 | Buildings' Shadow | 111 | 2122 | 2233 |
| 8 | Buildings | 312 | 5928 | 6240 |
| 9 | Sidewalk | 69 | 1316 | 1385 |
| 10 | Yellow Curb | 9 | 174 | 183 |
| 11 | Cloth Panels | 13 | 256 | 269 |
|  | Total | 2681 | 51006 | 53687 |

The dataset is 5×5 patches of high-spectral data with 53687 pixels as the center, and the pixels at the boundary are filled into 5×5 patches by interpolation. In all experiments in this section, 5% of the data of each category are randomly selected for training, and the remaining 95% are tested. The segmentation of the dataset is shown in Table 1.

(b) *Houston 2013 Dataset*: The Houston dataset was obtained over the University of Houston campus and adjacent urban areas using the ITRES-CASI 1500 sensor. The dataset has a spatial resolution of 2.5m and contains image with 144 spectral bands. The size of the image is 349×1905 pixels.

Table 2. Training and testing samples of Houston2013 Dataset, with 5% of the per class for training

| No. | Class | Train | Test | All |
|---|---|---|---|---|
| 1 | Healthy grass | 62 | 1189 | 1251 |
| 2 | Stressed grass | 62 | 1192 | 1254 |
| 3 | Synthetic grass | 34 | 663 | 697 |
| 4 | Trees | 62 | 1182 | 1244 |
| 5 | Soil | 62 | 1180 | 1242 |
| 6 | Water | 16 | 309 | 325 |
| 7 | Residential | 63 | 1205 | 1268 |
| 8 | Commercial | 62 | 1182 | 1244 |
| 9 | Roads | 62 | 1190 | 1252 |
| 10 | Highways | 61 | 1166 | 1227 |
| 11 | Railroads | 61 | 1174 | 1235 |

| 12 | Parking lot 1 | 61 | 1172 | 1233 |
| 13 | Parking lot 2 | 23 | 446 | 469 |
| 14 | Tennis court | 21 | 407 | 428 |
| 15 | Running track | 33 | 627 | 660 |
| | Total | 745 | 14284 | 15029 |

In particular, this HSI has a LiDAR derived Digital Surface Model (DSM) for the corresponding spatial resolution, which can further add features to the spectral information by combining spatial location and elevation information. In order to investigate the semantic understanding of the spectral model for better integration of spatial location and spectral information, the second dataset of this paper is a 5×5 patch of HSI centered on 15,029 pixels. The specific data segmentation is shown in Table 2.

**4.2 Comparative Methods**

To evaluate the network performance for unsupervised semantic understanding, this paper adopts the following comparative experiments on small-patch HSI and LiDAR fusion data:

(a) *Origin Data with SVM Classifier*: The data set $X$ fused by HSI and LiDAR data is directly classified by SVM without preprocessing, which is regarded as the baseline of other methods.

(b) *Principle Component Analysis (PCA)*: It can linearly reduce the dimension of high-dimensional data, and extract the features of main components after adjusting the dimensions of the original data for SVM classification.

(c) *Laplacian Eigenmaps (LE)*: It can locally understand the relationship between the data. This method retains the local linear structure of the data while reducing the dimension. After adjusting the dimension of the original data, it extracts the features of the original data that retain the local structure for SVM classification.

(d) *Cascade Residual Capsule Network* [30] *(CRCN)*: CRCN extracts high-dimensional data features by using residual and capsule network structures while considering the invariance features among HSI patches. Meanwhile, coding rate reduction (CRR) is introduced to measure the compactness of spectral features in the learning space. The difference between the encoding rate of the feature and the feature of each category is set as the loss function, and the feature with invariance is extracted unsupervised by reducing the loss. Then SVM classification is carried out.

(e) *SCAE* [31]: SCAE realizes unsupervised feature extraction of the spectrum based on the stacked capsule autoencoders model. First, it obtains parts with their pose by ConvLSTM, which extracts and transfers image features to Part Capsule. Meanwhile, it obtains the maximum likelihood of the input Part Capsule by Gaussian mixture model (GMM) decoding. Then, Set Transformer is used to extract the features of Part Capsule, and the description of HSI is available through these object capsules. Afterwards, the maximum likelihood of Part Capsule is gained through GMM. Finally, the unsupervised feature extraction model of HSI image is constructed.

(f) *ShearSAF* [32]: It is a shearlet-based structure-aware filtering approach, which can combine the texture and area information of HSI and LiDAR. It extracts the features of HSI by adopting kernel principal component analysis (KPCA) and transforms them into the shearlet domain after being merged with LiDAR. In this way, texture and regional features are obtained. Finally, the local adaptive filter is used to extract features, which are further classified by SVM.

**4.3 Experiment Settings**

The codes are made with the torch1.9.0 in the environment of python 3.7 and run in AMD7763 2.6GHz and NVIDIA A100. The learning rate is set as 0.001; the size of patch is 5×5; the optimizer is Adam; the high-dimensional canonical capsule is set to contain 15 capsules and each point has 50 channels. In order to verify the effectiveness of the network, the HSI and LiDAR features extracted from the network are classified by SVM on MATLAB.

To improve that the high-dimensional canonical capsule model can be effective for semantic understanding of HSI and LiDAR data, we carefully consider various types of accuracy, with average accuracy (AA), overall accuracy (OA) and Kappa coefficient in general.

**4.4 Experiment Results**

To illustrate the effectiveness of the high-dimensional canonical capsule for extracting features from hyperspectral data, we compare the classification results of extracted features containing semantic information with the original data using a support vector machine.

Table 3. Classification OA, AA, Kappa Results [%] of MUUFL

|  | BaseLine | | | PCA | LE | CRCN | SCAE | ShearSAF | Proposed |
|---|---|---|---|---|---|---|---|---|---|
|  | H | L | H+L | H+L | H+L | H | H | H+L | H+L |
| class1 | 91.71 | 94.08 | 96.41 | 96.19 | 96.39 | 96.14 | 96.69 | 97.36 | 96.59 |
| class2 | 48.17 | 69.97 | 67.07 | 58.95 | 52.91 | 73.29 | 76.87 | 79.76 | 76.48 |
| class3 | 58.35 | 80.80 | 78.03 | 81.52 | 81.18 | 84.10 | 80.08 | 79.52 | 84.50 |
| class4 | 24.61 | 77.51 | 81.23 | 74.34 | 73.53 | 84.42 | 85.93 | 85.41 | 88.29 |
| class5 | 61.97 | 89.81 | 93.29 | 81.41 | 82.65 | 92.44 | 93.40 | 94.02 | 92.74 |
| class6 | 92.21 | 79.86 | 8.52 | 13.35 | 3.62 | 82.93 | 84.39 | 78.05 | 82.81 |
| class7 | 5.97 | 66.38 | 66.83 | 9.24 | 9.48 | 73.37 | 72.04 | 64.03 | 76.52 |
| class8 | 63.20 | 84.56 | 88.48 | 74.46 | 67.95 | 91.99 | 93.62 | 94.79 | 93.54 |
| class9 | 9.28 | 43.35 | 53.13 | 9.20 | 13.00 | 48.41 | 47.15 | 33.76 | 57.19 |
| class10 | 2.98 | 31.79 | 4.76 | 76.30 | 67.05 | 33.93 | 47.40 | 24.86 | 31.79 |
| class11 | 5.10 | 89.02 | 81.18 | 94.12 | 94.90 | 94.14 | 90.20 | 85.88 | 89.80 |
| OA | 68.33 | 85.44 | 86.44 | 79.58 | 78.51 | 88.93 | 89.24 | 89.05 | **90.10** |
| AA | 42.14 | 73.38 | 65.36 | 60.83 | 58.42 | 77.74 | 78.89 | 74.31 | **79.11** |
| kappa | 56.18 | 80.68 | 81.85 | 72.05 | 70.57 | 85.29 | 85.72 | 85.39 | **86.87** |

According to the above experimental setup, we verified the effect of fusion of HSI and LiDAR in the designed network model, and MUUFL results are shown in Table 3. According to the results in Table 3, the classification effect of the tenth category is barely satisfactory due to the scarce number of categories. However, it can be seen that the model proposed in this paper has the highest OA performance. The feature extraction effect of the model with semantic understanding proposed in this paper is much better than other feature extraction methods such as PCA and LE, and slightly better than SCAE. Due to data imbalance, the accuracy of categories with fewer samples decreases significantly after adding spatial information, while the accuracy of categories with more samples increases slightly after adding spatial information. In terms of AA performance, the model HSI+LiDAR proposed in this paper has better effect than HSI.

Table 4. Classification OA, AA, Kappa Results [%] of Houston2013

|  | BaseLine | | | PCA | LE | CRCN | SCAE | ShearSAF | Proposed |
|---|---|---|---|---|---|---|---|---|---|
|  | H | L | H+L | H+L | H+L | H | H | H+L | H+L |
| class1 | 98.49 | 96.19 | 96.39 | 91.75 | 88.13 | 85.95 | 86.54 | 84.93 | 93.10 |
| class2 | 88.76 | 49.92 | 96.14 | 90.09 | 95.38 | 99.83 | 99.41 | 96.81 | 98.91 |
| class3 | 90.80 | 63.35 | 92.76 | 100.00 | 100.00 | 98.49 | 97.89 | 99.70 | 99.40 |
| class4 | 91.20 | 81.47 | 97.38 | 97.54 | 95.51 | 93.49 | 99.07 | 98.48 | 98.39 |
| class5 | 100.00 | 55.76 | 100.00 | 98.90 | 99.15 | 97.46 | 98.22 | 98.98 | 98.81  |
| class6 | 66.02  | 8.09  | 73.79  | 97.73 | 97.73 | 93.20 | 83.82 | 83.77 | 82.52  |
| class7 | 84.65  | 63.90 | 89.13  | 85.96 | 83.89 | 94.94 | 92.37 | 91.61 | 96.60  |
| class8 | 70.56  | 80.29 | 73.10  | 74.26 | 68.50 | 81.39 | 87.31 | 88.82 | 92.22  |
| class9 | 83.11  | 34.87 | 86.05  | 52.23 | 45.25 | 80.25 | 87.31 | 80.66 | 87.31  |
| class10| 82.33  | 47.94 | 92.54  | 49.27 | 40.86 | 97.43 | 97.43 | 81.97 | 98.37  |
| class11| 78.62  | 76.83 | 82.79  | 52.43 | 38.02 | 95.32 | 93.10 | 80.73 | 95.74  |
| class12| 89.33  | 43.43 | 81.40  | 46.37 | 28.52 | 82.85 | 76.71 | 88.56 | 87.20  |
| class13| 88.34  | 44.62 | 83.63  | 13.48 | 6.52  | 26.91 | 32.51 | 83.60 | 23.32  |
| class14| 89.43  | 13.27 | 91.65  | 84.24 | 80.30 | 96.31 | 97.05 | 98.77 | 100.00 |
| class15| 98.72  | 71.45 | 99.52  | 99.04 | 99.20 | 98.56 | 99.36 | 99.20 | 99.68  |
| OA     | 87.11  | 57.08 | 89.60  | 75.21 | 70.31 | 89.79 | 90.50 | 90.08 | **92.77** |
| AA     | 86.69  | 52.87 | 89.06  | 75.55 | 71.13 | 88.16 | 88.54 | **90.44** | 90.11 |
| kappa  | 86.07  | 53.50 | 88.75  | 73.16 | 67.85 | 88.95 | 89.72 | 89.28 | **92.17** |

From the results of Houston shown in Table 4, it is implied that our high-dimensional canonical capsule has the ability to effectively extract HSI+LiDAR features. In most cases, the models using both HSI+LiDAR perform better than the models using only HSI. Besides, the model we proposed is superior to the ShearSAF model by 2.69%. The unsupervised effect of HSI and LiDAR fused is better than that of baseline and far better than that of linear feature extraction methods like PCA and LE. Therefore, based on the semantic understanding within the data, it can uncover some connections between spatial and spectral features to achieve better results than the original data. It can be seen that the fused semantics can more effectively extract spectral features, but from the respect of accuracy, the model proposed in this paper is less effective in classifying parking lot 2, which also indicates that there are some omissions in the model's understanding of spectral semantics. In general, the model proposed in this paper, which combined spectral semantic information and spatial location, can effectively extract the features of HSI.

## 5. Conclusions and future work

In this paper, we present an unsupervised network model for spectral-spatial-elevation fusion feature extraction and semantically understand the spatial and spectral information from invariance and equivariance of the network. Extracting semantic information of complex multi-source RS data and revealing superior performances in unsupervised feature extraction as opposed to SCAE, it can be concluded that the algorithms in current study have promising applications in feature extraction of remote sensing data.